\documentclass{article}

\usepackage{arxiv}

\usepackage{wrapfig}
\usepackage[table,dvipsnames]{xcolor}
\usepackage[utf8]{inputenc} 
\usepackage[T1]{fontenc}    
\usepackage{hyperref}       
\usepackage{url}            
\usepackage{booktabs}       
\usepackage{amsmath,amssymb,amsfonts}       
\usepackage{nicefrac}       
\usepackage{microtype}      
\usepackage{graphicx}
\usepackage{natbib}
\usepackage{doi}
\usepackage{multirow}
\usepackage{cleveref}
\usepackage{soul}
\usepackage{pifont}
\usepackage{gensymb}
\usepackage{mathtools}
\usepackage{comment}

\newcommand{\result}[2]{\ensuremath{#1}\scriptsize{$\pm$\ensuremath{#2}}}

\hypersetup{
    colorlinks,
    linkcolor={red!80!black},
    citecolor={blue!50!black},
    urlcolor={blue!80!black}
}

\title{Evidential Federated Learning for Skin Lesion Image Classification}

\author{ \href{https://orcid.org/0009-0002-9492-5954}{\includegraphics[scale=0.06]{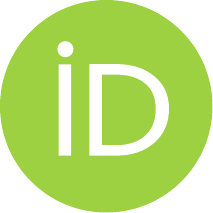}\hspace{1mm}Rutger Hendrix} \\
	Department of Electrical, Electronic\\
    and Computer Engineering\\
	University of Catania, Italy\\
	\texttt{rutger.hendrix@phd.unict.it} \\
	\And
	\href{https://orcid.org/0000-0002-6122-4249}{\includegraphics[scale=0.06]{orcid.pdf}\hspace{1mm}Federica Proietto Salanitri} \\
	Department of Electrical, Electronic\\
    and Computer Engineering\\
	University of Catania, Italy\\
	\texttt{federica.proiettosalanitri@unict.it} \\
    \And
	\href{https://orcid.org/0000-0001-6653-2577}{\includegraphics[scale=0.06]{orcid.pdf}\hspace{1mm}Concetto Spampinato} \\
	Department of Electrical, Electronic\\
    and Computer Engineering\\
	University of Catania, Italy\\
	\texttt{concetto.spampinato@unict.it} \\
    \And
	\href{https://orcid.org/0000-0002-2441-0982}{\includegraphics[scale=0.06]{orcid.pdf}\hspace{1mm}Simone Palazzo} \\
	Department of Electrical, Electronic\\
    and Computer Engineering\\
	University of Catania, Italy\\
	\texttt{simone.palazzo@unict.it} \\
    \And
	\href{https://orcid.org/0000-0001-7379-6829}{\includegraphics[scale=0.06]{orcid.pdf}\hspace{1mm}Ulas Bagci} \\
	Machine \& Hybrid Intelligence Lab\\
    Department of Radiology\\
	Northwestern University, USA\\
	\texttt{ulas.bagci@northwestern.edu} \\
}

\date{}

\renewcommand{\headeright}{\raggedright Published as a conference paper at ICPR 2024}
\renewcommand{\undertitle}{}
\renewcommand{\shorttitle}{}

\hypersetup{
pdftitle={FedEvPrompt},
pdfsubject={Machine Learning: Federated Learning, Prompt Tuning, and Knowledge Distillation in Medical Applications},
pdfauthor={R. Hendrix, F. Proietto Salanitri, C. Spampinato, S. Palazzo, U. Bagci},
pdfkeywords={Prompt Tuning, Knowledge Distillation, Uncertainty},
}

\begin{document}
\maketitle

\begin{abstract}
	We introduce \textit{FedEvPrompt}, a federated learning approach that integrates principles of evidential deep learning, prompt tuning, and knowledge distillation for distributed skin lesion classification. \\ FedEvPrompt leverages two sets of prompts: \textit{b-prompts} (for low-level basic visual knowledge) and \textit{t-prompts} (for task-specific knowledge)  prepended to frozen pre-trained Vision Transformer (ViT) models trained in an evidential learning framework to maximize class evidences. Crucially, knowledge sharing across federation clients is achieved only through knowledge distillation on attention maps generated by the local ViT models, ensuring enhanced privacy preservation compared to traditional parameter or synthetic image sharing methodologies. FedEvPrompt is optimized within a round-based learning paradigm, where each round involves training local models followed by attention maps sharing with all federation clients.
Experimental validation conducted  in a real distributed setting, on the ISIC2019 dataset, demonstrates the superior performance of FedEvPrompt against baseline federated learning algorithms and knowledge distillation methods, without sharing model parameters. In conclusion, FedEvPrompt offers a promising approach for federated learning, effectively addressing challenges such as data heterogeneity, imbalance, privacy preservation, and knowledge sharing.
\end{abstract}


\section{Introduction}
\label{sec:intro}
In recent decades, deep learning has played a leading role in medical image analysis, including skin lesion classification. However, most of the existing methods rely on centralized learning, assuming data uniformity and accessibility, which often does not align with the reality of decentralized and privacy-sensitive clinical settings. This disparity not only limits progress in the field, but also exacerbates inequalities, with wealthier regions having a data advantage over poorer areas, leading to disparities in model performance and clinical support. Federated learning (FL) emerges as a promising solution to this challenge, enabling model training across distributed devices while preserving data privacy. Methods like FedAvg~\cite{mcmahan2017communication} and FedProx~\cite{li2020federated} have addressed issues such as non-i.i.d. data and system heterogeneity, yet they still face obstacles, particularly in scenarios with class imbalances and data heterogeneity.
Evidential Deep Learning (EDL)~\cite{sensoy2018evidential} has found adoption in FL to handle these limitations in medical data, thereby enhancing model confidence and reliability, crucial for clinical applications. For example, the recent work on uncertainty-aware  aggregation of federated models for diabetic retinopathy classification demonstrates its efficacy in improving model performance and reliability~\cite{wang2023federated}.\\
Furthermore, the scarcity of data poses an additional significant challenge, often leading to model overfitting and suboptimal federation performance. Recent techniques like learnable prompting~\cite{li2021prefix}, particularly effective in low-data regimes, offer a promising solution by facilitating personalized model tuning across distributed clients~\cite{li2023visual}.
Nonetheless, privacy concerns persist, particularly due to the sharing and aggregation of model parameters, which poses the risk of reconstructing training images, as demonstrated by recent studies~\cite{zhang20233d,geiping2020inverting,zhu2019deep}.
To mitigate these concerns, one strategy involves sharing suitably-constructed synthetic data generated through generative models~\cite{pennisi2024feder}. Yet, the use of generative models carries its own risks, potentially incorporating and synthesizing sensitive training samples, thus exacerbating privacy concerns.\\
We here propose FedEvPrompt, a novel approach that integrates principles of evidential deep learning, prompt tuning, and knowledge distillation to address existing limitations comprehensively. FedEvPrompt leverages prompts prepended to pre-trained ViT models trained in an evidential learning setting, maximizing class evidence. Knowledge sharing across federation clients is achieved only through knowledge distillation on attention maps generated by ViT models, which offers greater privacy preservation compared to sharing parameters or synthetic images, as it lacks pixel-level details and reconstructive qualities. While our approach maintains a high level of abstraction for minimizing privacy leaks, it also provides richer information than average logits, as in FedDistill~\cite{seo202216}, or prototypes, as in FedProto~\cite{tan2022fedproto}.
Thus, FedEvPrompt represents a principled way to share insights into the decision-making process of local models for enhanced federated performance, as demonstrated by the results achieved on a real-word distributed setting for skin lesion classification.

\section{Background Evidential Learning}
\label{sec:background}

Deep Learning methods often use softmax activation in the output layers to perform classification. However, softmax outputs can be biased to training data, failing to predict with low certainty even for samples far from the distribution \cite{staahl2020evaluation}. In contrast to the additivity principle in probability theory, Dempster-Shafer theory describes that the sum of belief can be less than 1. Its remainder is then attributed to uncertainty. 

In a frame of \(K\) mutually exclusive singletons (e.g., class labels), each singleton \(k \in [K]\) is assigned a belief mass \(b_k\), and an overall uncertainty mass \(u\). The sum of these \(K + 1\) mass values is constrained by \(u + \sum_{k=1}^K b_k = 1\), with \(u \geq 0\), \(b_k \geq 0\), \(\forall k \in [K]\). The belief mass is determined by the evidence supporting each singleton, reflecting the level of support gathered from data. The uncertainty is inversely proportional to the total amount of evidence, with uncertainty equal to 1 for a total lack of evidence. A belief mass assignment corresponds to a Dirichlet distribution with  parameters \(\alpha_k = e_k + 1\), where \(e_k\) denotes the derived evidence for the \(k\)-th singleton. 
This choice of Dirichlet distribution is motivated by its role as a conjugate prior to the categorical distribution, and is defined as:

\[
\text{Dir}(p, \alpha) = \frac{\Gamma(S)}{\prod_{k=1}^K \Gamma(\alpha_k)} \prod_{k=1}^K p_k^{\alpha_k - 1}, \quad \alpha_k > 0
\]

where \(p\) denotes a probability mass function, \(K\) denotes the number of classes, \(\alpha = [\alpha_1, \ldots, \alpha_K]\) are the Dirichlet parameters related to the evidence, \(\Gamma(\cdot)\) denotes the gamma function, and \(S = \sum_{k=1}^K \alpha_k\) is termed the Dirichlet strength.

From the parameters of this Dirichlet distribution, the belief \(b_k\) and the uncertainty \(u\) are derived as:

\[
b_k = \frac{\alpha_k - 1}{S}, \quad u = \frac{K}{S}
\]

When considering an opinion, the expected probability \(\hat{p}_k\) of the \(k\)-th singleton equates to the mean of the corresponding Dirichlet distribution, calculated by:

\[
\hat{p}_k = \frac{\alpha_k}{S} 
\]

Although this modeling of second-order probabilities and uncertainty enables the computation of different types of uncertainties, this work  only considers classical vacuity uncertainty (\(u\)).

Evidential Deep Learning (EDL) aims to quantify these uncertainties in the predictions, using a single deterministic neural network. The model learns evidence from the logit layer, typically applying non-negative functions like ReLU to obtain these values. With these minimal changes, EDL models can be trained by minimizing losses such as evidential mean squared error (MSE) loss to form the multinomial opinions for \( K \)-class classification of a given sample \( i \) as a Dirichlet distribution. Following ~\cite{NEURIPS2018_a981f2b7}, the evidential MSE loss for sample i can be interpreted as:
\[
\mathcal{L}_{i}(\Theta) = \int \left[(\mathbf{y}_i - \mathbf{p}_i)^T (\mathbf{y}_i - \mathbf{p}_i)\right]\text{Dir}(\mathbf{p}_i, \mathbf{\alpha}_i) d \mathbf{p}_i
\]
\[
\mathcal = \sum_{k=1}^{K} \mathbb{E}\left[y_{i,k}^2 - 2 y_{i,k} y_{i,k} + p_{i,k}^2\right]
\]

where \( \mathbf{y}_i = (y_{i,1}, \ldots, y_{i,K}) \) and \( \mathbf{p}_i = (p_{i,1}, \ldots, p_{i,K}) \) are vectors of true and predicted probabilities, and \( \alpha_i = (\alpha_{i,1}, \ldots, \alpha_{i,K}) \) are the Dirichlet parameters.

Using the identity \( \mathbb{E}\left[p_{i,k}^2\right] = \mathbb{E}\left[p_{i,k}\right]^2 + \operatorname{Var}(p_{i,k}) \), the loss can be rewritten as:
\[
\mathcal{L}_{i}(\Theta) = \sum_{k=1}^{K} \left( (y_{i,k} - \mathbb{E}\left[p_{i,k}\right])^2 + \operatorname{Var}(p_{i,k}) \right)
\]
\[
 = \sum_{k=1}^{K} \left( \left(y_{i,k} - \frac{\alpha_{i,k}}{S_k}\right)^2 + \frac{\alpha_{i,k} \left(S_i - \alpha_{i,k}\right)}{S_i^2 (S_i + 1)} \right)
\]
where \( S_i \) is the total Dirichlet strength for sample \( i \).

To avoid generating misleading evidence for incorrect labels, a Kullback-Leibler (KL) divergence regularization term is used, reducing total evidence to zero for incorrectly classified samples. The KL term is defined as:
\[
\mathcal{L}_{KL} = \text{KL} \left[ \text{Dir}(p_i \mid \widetilde{\boldsymbol{\alpha}}_i) \| \text{Dir}(p_i \mid \mathbf{1}) \right] 
\]
\[
\quad = \log \left( \frac{\Gamma\left(\sum_{k=1}^{K} \widetilde{\alpha}_{i,k}\right)}{\Gamma(K) \prod_{k=1}^{K} \Gamma(\widetilde{\alpha}_{i,k})} \right) + \sum_{k=1}^{K} \left( \widetilde{\alpha}_{i,k} - 1 \right) \left( \psi(\widetilde{\alpha}_{i,k}) - \psi\left(\sum_{k=1}^{K} \widetilde{\alpha}_{i,j}\right) \right)
\]
where \( \widetilde{\boldsymbol{\alpha}}_i \) are the Dirichlet parameters after the removal of non-misleading evidence defined as $\widetilde{\boldsymbol{\alpha}} = y  + (1 - y) \odot \alpha$,
\(\text{KL}[. \| .]\) denotes the Kullback-Leibler divergence operator, and \(\psi(.)\) is the digamma function ~\cite{NEURIPS2018_a981f2b7}.
The final \textbf{evidential loss} $\mathcal{L_\epsilon}$ results in:
\[
\mathcal{L_\epsilon} = \mathcal{L}_{i} + \mathcal{L}_{KL}
\]

\section{Methodology}
\label{sec:method}
\begin{figure}
    \centering
    \includegraphics[width=\textwidth]{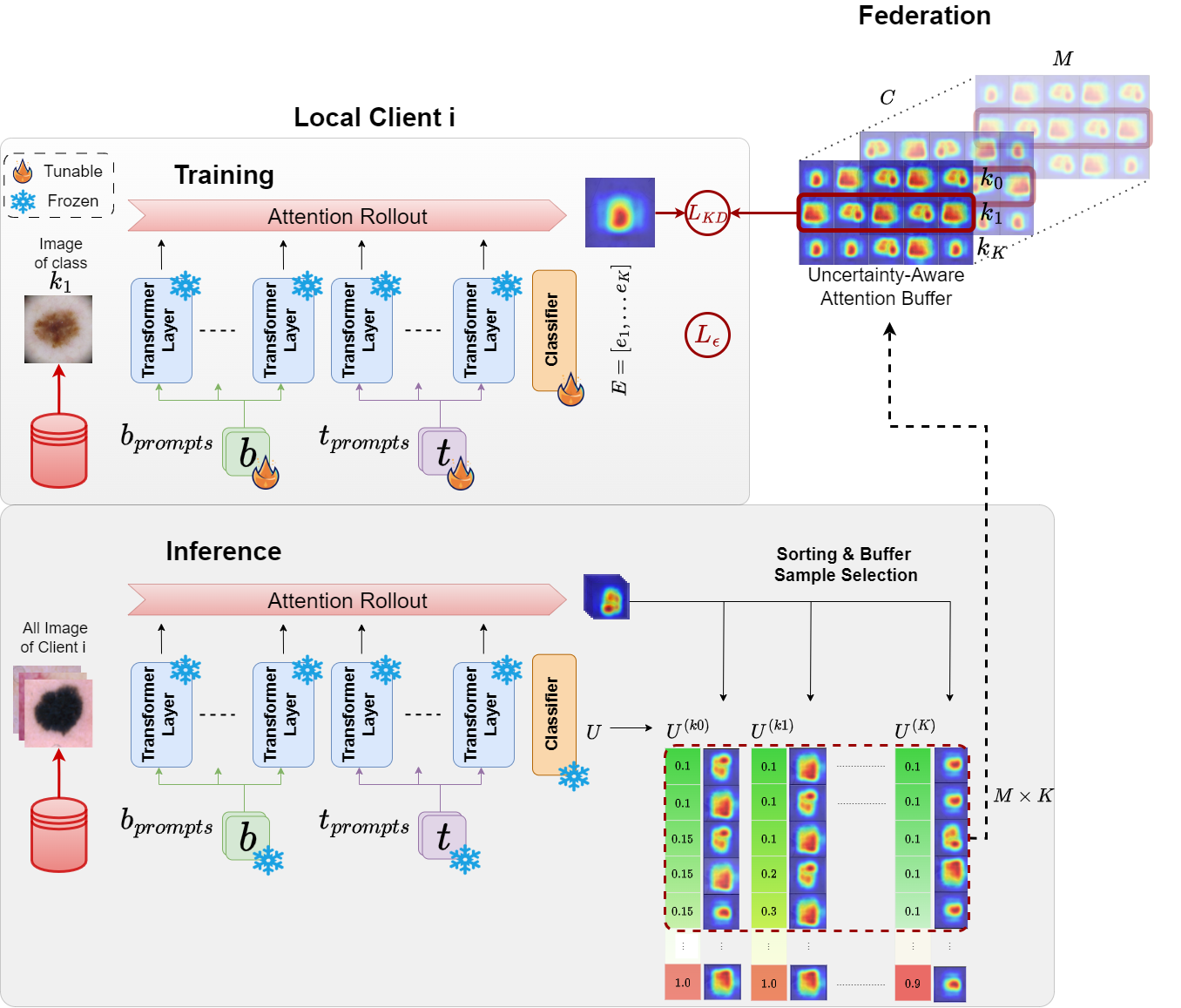}
    \caption{    \textbf{Overview of FedEvPrompt}. During a round of \textbf{Training (top)}, local data is used to optimize \textit{b-prompts}, encoding general visual features, and \textit{t-prompts}, encoding task-specific information, prepended to a frozen ViT encoder. Optimization is carried out by minimizing evidential loss ($\mathcal{L_{\epsilon}}$) and a knowledge distillation loss ($\mathcal{L_{KD}}$) between local attention maps and those of the federation available in the \textit{uncertainty-aware attention buffer}. After a round of training at \textbf{Inference (bottom)}, the client identifies, for each class $K$, its $M$ most informative attention rollout maps (sorted by lowest uncertainty) to contribute to the federated \textit{uncertainty-aware attention buffer}.}
    \label{fig:strategy}
\end{figure}

We introduce FedEvPrompt, our federated learning paradigm, which leverages prompt evidential learning and knowledge distillation on ViT attention maps for enabling effective knowledge aggregation across federated clients. The overall learning strategy is described in Fig.~\ref{fig:strategy}. 

FedEvPrompt is based on a pre-trained ViT model, kept frozen across all clients within the federation. Upon the fixed backbone, prompts are prepended on each client model and optimized using local data. Each client also computes attention maps  (through attention rollout mechanism~\cite{abnar2020quantifying}) for each class and shares a subset of them with the federation. The attention maps by all clients form our \textit{uncertainty-aware attention buffer} that is used for knowledge distillation during prompt learning. 

Learning is organized in rounds: at each round, federation clients carry out different local training epochs for prompt optimization through a combination of evidential loss for learning class evidence and knowledge distillation loss on the per-class attention maps present within the buffer.
At the end of training round, each client identifies its $M$ most informative attention maps for each class and updates the buffer.\\  More in detail, each client employs a frozen ViT, as the backbone, with two sets of prompts \textit{b-prompts} and \textit{t-prompts}.  Each prompt is associated with a specific attention layer, with the \textit{b-prompts} (basic prompts) prepended to layers with low-level feature representation and the \textit{t-prompts} (task-specifc prompts) to deeper layers with high-level feature representation. The parameters of the prompts are incorporated through pre-fix tuning.
Let's denote the output of the $i_{th}$ attention layer as $h_i$ with $i \in 1 \cdots H$ where $H$ is the number of attention layers. The prepended parameters for the key and value inputs, denoted as $pr_k$ and $pr_v$ respectively, are introduced as follows:
\begin{equation}
\label{eq:msa}
{MSA}(h_Q, [pr_k^{(i)}; h_K], [pr_v^{(i)}; h_V])
\end{equation}

where $h_Q$, $h_K$, and $h_V$ represent the query, key, and value outputs from the previous layer, respectively. 
The prepended prompts $pr_k^{(i)}$ and $pr_v^{(i)}$ are \textit{b-prompts} to the first $l$ layers, and \textit{t-prompts} to the last $H-l$ layers.
The two sets of prompts undergo distinct optimization strategies: \textit{b-prompts} require slower adaptation since the frozen backbone (ViT) has already grasped general visual features. Conversely, \textit{t-prompts} necessitate faster adjustments to accommodate varying data distributions. Consequently, $\mu_1 < \mu_2$, where $\mu_1$ and $\mu_2$ denote the learning rates for the \textit{b-prompts} and \textit{g-prompts} optimizers, respectively. Both set of prompts are optimized by minimizing an overall loss $\mathcal{L}_{G}$ that includes an evidential loss term $\mathcal{L_\epsilon}$ and a knowledge distillation loss term $\mathcal{L_{KD}}$ on the shared attention map buffer $A$:
\begin{equation}
    \mathcal{L}_{G}  = \mathcal{L_\epsilon} + \lambda \mathcal{L_{KD}}
\end{equation}
where $\lambda=1e^{-6}$ is a parameter controlling the balance between the two terms.

\subsection{Evidential loss}
\label{sec:ev}
Our method is based on evidential learning, i.e., the classification model outputs evidences $E = [e_1, \ldots, e_K]$, with $K$ categorical class elements (number of classes).
The Dirichlet distribution characterizes the likelihood of each discrete probability value within a set of possible probabilities. It is parameterized by a vector of $K$ elements (classes), $\boldsymbol{\alpha} = [\alpha_1, \ldots, \alpha_K]$, defined as  $\alpha_k = e_k + W_k$, 
with $e_k$ being the model evidence for class $k$, and $W_k$ the prior weight for that class. Classical EDL assumes a uniform Dirichlet ($\text{Dir}(1)$) distribution as a prior, i.e., $\boldsymbol{W} = \langle 1,1,\ldots,1 \rangle$. The uncertainty for the $i^{th}$ input sample is then estimated as $u_i = \frac{K}{S}$, with $S$ being the total Dirichlet strength $S = \sum_{k=1}^{K}\alpha_k$.
Due to the strong class imbalance typical in federated learning settings, we change the uniform evidential prior to a skewed distribution, weighted by class frequency: 
\begin{equation}
\alpha_k=e_k+W_k \quad \text{with} \quad W_k=\frac{K}{K-1}\left(1-\frac{N_k}{N}\right)
\end{equation}
such that $\sum_{k=1}^{K}W_k = K$.

Prompt parameters are finally optimized by minimizing the evidential loss defined in~\cite{NEURIPS2018_a981f2b7}  as a combination of MSE and KL divergence. 
Given the $i^{th}$ input sample, the one-hot-encoded vector  $y_i$ of its class label $k$, and its expected probabilities $p_i$, the \textbf{evidential loss} $\mathcal{L_\epsilon}$ is computed as:
\begin{equation}
\label{eq:ev_loss}
\mathcal{L_\epsilon}(\boldsymbol{\theta}) = \mathbb{E}_{\mathbf{p} \sim \text{Dir}(\boldsymbol{\alpha})}\left[(y_i - p_i)^T (y_i - p_i)\right] + \lambda_{KL} D_{KL}(\text{Dir}(p_i | \widetilde{\alpha}_i) || \text{Dir}(p_i | \mathbf{w}_i))
\end{equation}
with $\lambda_{KL} = min(1,t/10)$ being an annealing factor applied to gradually increase the regularization impact with the number of epochs $t$. 

In order to let the evidence for incorrect classes shrink to the weighted prior values $\mathbf{W}$, the KL divergence loss term minimizes the distributional difference between $\mathbf{W}$ and misleading evidence $\widetilde{\boldsymbol{\alpha}}$, formulated as $\widetilde{\boldsymbol{\alpha}} = y \cdot \mathbf{w}  + (1 - y) \odot \boldsymbol{\alpha}$. Given the weighted prior distribution $\mathbf{W} = \text{Dir}(\mathbf{p} | \mathbf{w})$ and $P = \text{Dir}(\mathbf{p} | \widetilde{\boldsymbol{\alpha}})$, the general KL divergence form for the Dirichlet distribution \cite{ProofPennyKL} becomes:

\begin{equation}
\begin{aligned}
& D_{KL}(\text{Dir}(p | \widetilde{\alpha}) || \text{Dir}(p | \mathbf{w})) = 
 \log\left(\frac{\Gamma\left(\sum_{k=1}^K \widetilde{\alpha}_k\right)}{\Gamma\left(\sum_{k=1}^K w_k\right)}\right) + \sum_{k=1}^K \log\left(\frac{\Gamma(w_k)}{\Gamma(\widetilde{\alpha}_k)}\right) + \\
 &\sum_{k=1}^K (\widetilde{\alpha}_k - w_k) \cdot \left[\psi(\widetilde{\alpha}_k) - \psi\left(\sum_{k=1}^K \widetilde{\alpha}_k\right)\right] = \log\left(\frac{\Gamma\left(\sum_{k=1}^K \widetilde{\alpha}_k\right) \cdot \prod_{k=1}^{K}\Gamma\left(w_k\right)}{\Gamma\left(K\right) \cdot \prod_{k=1}^{K}\Gamma\left(\widetilde{\alpha}_{k}\right)}\right) + \\
 & +  \sum_{k=1}^K (\widetilde{\alpha}_{k} - w_{k}) \cdot \left[\psi(\widetilde{\alpha}_{k}) - \psi\left(\sum_{k=1}^K \widetilde{\alpha}_k\right)\right]
\end{aligned}
\end{equation}

With $\Gamma$ being the gamma function, and $\psi$ being the digamma function.\\

\subsection{Uncertainty-Aware Attention Buffer for Knowledge Distillation}
\label{sec:kd}

Prompt optimization involves minimizing a knowledge distillation loss $\mathcal{L}_{KD}$ term between the model attention maps (computed through attention rollout) and the maps available in our \textit{uncertainty-aware attention buffer}  \( A \) shared within all the $C$ clients of the federation, with each client providing $M$ attention maps for each of the $K$ classes:

\begin{equation}
\label{eq:buffer}
A = \bigcup_{c=1}^{C} \bigcup_{k=1}^{K} \bigcup_{m=1}^{M} a_{c,k,m}    
\end{equation}

here $a_{c,k,\textbf{$i$}} \in \mathcal{R}^{H\times W}$ represents the $i^{th}$ attention map for the $k^{th}$ class of the $c^{th}$ client of the federation. $H$ and $W$ are the height and width of the attention maps equal to input image dimensions. The \textbf{knowledge distillation loss} $\mathcal{L}_{KD}$ for a generic training sample of client $c$, with class $k$ can be expressed as:

\begin{equation}
\mathcal{L}_{KD}  =   \frac{1}{M} \sum_{i=1}^{C\setminus c} \sum_{m=1}^{M} \left\| a_{c,k,\_} - a_{i,k,m} \right\|^2
\end{equation}

where, $\left\| a_{c,k,\_} - a_{i,k,m} \right\|^2$ denotes the squared Euclidean distance between the attention map $a_{c,k,\_}$ of the considered training sample and $a_{i,k,m}$ being an item of the buffer $A$. \\

The selection of samples for the attention buffer $A$ by each client is based on the assumption that each local model should share its most confident predictions and indicate the image regions it focuses on. We use uncertainty scores from our evidential learning approach to guide the selection of attention maps for sharing within the federation. Specifically, we compute uncertainty scores, $u_{j}^{(k)}$, for each sample in class $k$, where $j$ ranges from 1 to $N^{(k)}$, the total number of samples in class $k$. From these, we select the $M$ samples with the lowest uncertainty scores, denoted as $\{u_{1}^{(k)}, u_{2}^{(k)}, \ldots, u_{M}^{(k)} \}$, and corresponding attention maps for inclusion in our uncertainty-aware attention buffer, $A$, replacing older ones.

\section{Experimental results}
\label{sec:experiment}
We validate the effectiveness of our proposed method on a multicenter dataset of 23,247 dermoscopic images of nine skin lesions from different populations and medical centers, based on the ISIC2019 dataset~\cite{combalia2019bcn20000,codella2018skin,tschandl2018ham10000}.  To carry out federated learning, we organized the dataset into six nodes, with each node representing data from a specific source: 
\begin{figure}[h!]
   \centering
    \includegraphics[width = 0.85\textwidth]{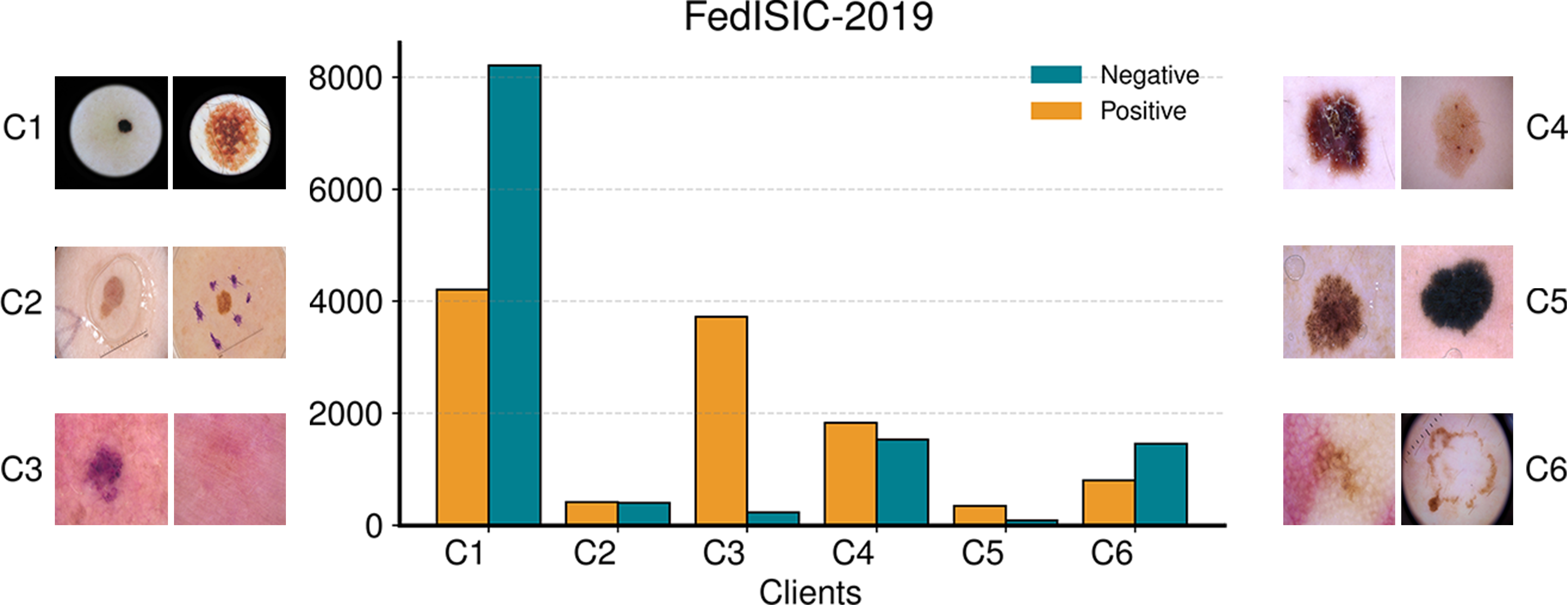}
    \caption{Distribution of the Fed-ISIC2019 dataset across clients.}
    \label{fig:dataset}
\end{figure}
Client C1 contains the BCN20000 dataset as described by Combalia et al.~\cite{combalia2019bcn20000}, which includes 19,424 images from the Hospital Clínic in Barcelona; Clients C2, C3 and C4 are from the Austrian portion of the HAM10000 dataset~\cite{tschandl2018ham10000}, with images from the ViDIR Group at the Department of Dermatology at the Medical University of Vienna; Client C5 is also part of the HAM10000 dataset and contains the Rosendahl image set from the University of Queensland in Australia; while client C6 includes the MSK4 dataset~\cite{codella2018skin}. 
The overall dataset exhibits heterogeneity in both the number of images contributed by each client as well ass in the distribution of classes, as illustrated in Fig.~\ref{fig:dataset}, making it a strong real-world use case for testing federated learning methods. For this study, we will focus on the binary classification task of distinguishing \textit{Melanocytic nevus} from other skin lesions.\\
\\
\textbf{Training procedure}
In our setup, for each client, data is divided into a 75\% training and 25\% test split. Training is executed over 5 communication rounds, with 15 training epochs per round. Our model architecture employs a frozen ViT backbone augmented with additional parameters for b-prompts, t-prompts, and a classification-head.
The ViT backbone specifications include an embedding dimension of 384, 6 attention heads, 12 blocks, and an input size of 224x224 pixels. For the b-prompts and t-prompts, the prompt keys ($k$ and $v$) have a sequence length of 50, while $l$ is 3 out of the 12 attention layers. We set the learning rates $\mu_1$ and $\mu_2$ to $2.5e-4$ and $5e-4$ respectively, with a weight decay factor of $1e-2$. Additionally, each client contributes 5 attention rollout maps per class (i.e., $M$ in Eq.~\ref{eq:buffer}) to the uncertainty-aware attention buffer. 
Results are presented as in terms of balanced accuracy on the test set at the conclusion of all rounds.\\
\textbf{Results}. In Table~\ref{tab:sota}, we present a comprehensive performance comparison between FedEvPrompt and existing federated learning methods. FedAvg~\cite{mcmahan2017communication} serves as our baseline. FedAvgPers builds upon FedAvg by integrating a personalization step through local data fine-tuning, aligning with our emphasis on personalized learning via b-prompts and t-prompts tuning. Additionally, we incorporate FedProx~\cite{li2020federated}, specifically tailored to address non-IID data like our skin lesion dataset.\\
Given that our approach employs knowledge distillation without parameter sharing, we include two analogous methods in our analysis: FedProto~\cite{tan2022fedproto} and FedDistill~\cite{seo202216}. 
We also evaluate the performance of local training, where client models are trained independently without parameter sharing, using both sets of prompts (i.e., $(b,t)_{prompts}$), and using only one set of prompts ($g_{prompts}$ - general prompts) across all attention layers. We define $\emph{g-prompts} = [\emph{b-prompts},\emph{t-prompts}]$ with both learning rates set to $\mu_{1}$.
This evaluation aims to validate our choice to apply different parameters across different attention layers and to demonstrate the advantages provided by our federated learning approach.\\
Results show that FedEvPrompt outperforms its competitors, including those that share parameters (thus being less privacy-preserving), such as FedAvg~\cite{mcmahan2017communication} and FedProx~\cite{li2020federated}. Notably, when comparing FedEvPrompt performance with other methods that do not share parameters, namely FedProto~\cite{tan2022fedproto} and FedDistill~\cite{seo202216}, we observe higher performance across all clients and a lower standard deviation, indicating better convergence in accuracy among clients.

\begin{table}
\centering
\caption{\textbf{Comparison with state-of-the-art methods} on the Skin Lesion Dataset. In bold, best accuracy values.}
\rowcolors{1}{white}{gray!10}
\label{tab:sota}
\footnotesize{
\begin{tabular}{r|cccccc|c} 
\toprule
                                            & \textbf{C1}   & \textbf{C2}    & \textbf{C3}   & \textbf{C4}   & \textbf{C5}   & \textbf{C6}  & \textbf{Avg}\\ 
\midrule
Local $b,t_{prompts}$                                   & ~~$72.56$     & ~~$50.00$      & ~~$89.91$     & ~~$79.94$     & ~~$67.61$     & ~~$50.00$    & ~~\result{$68.34$}{$10.98$}      \\
Local  $g_{prompts}$               & ~~$50.00$     & ~~$50.48$    & ~~$84.79$    & ~~$78.91$    & ~~$71.59$    & ~~$50.00$    & ~~\result{$64.30$}{$10.31$}\\

\midrule
FedAvg$$[\textcolor{violet}{3}]$$& ~~$74.74$     & ~~$72.79$    & ~~$67.74$    & ~~$79.89$    & ~~$67.61$    & ~~$77.09$    & ~~\result{$73.31$}{$3.64$}\\
FedAvgPers                                  & ~~$77.79$     & ~~$68.01$      & ~~$84.84$     & ~~$78.08$     & ~~$67.61$     & ~~$82.66$    & ~~\result{$76.50$}{$7.25$}       \\
FedProx$$[\textcolor{violet}{4}]$$               & ~~$81.34$     & ~~$68.14$      & ~~$74.69$     & ~~$75.15$     & ~~$77.84$     & ~~$78.65$    & ~~\result{$75.97$}{$4.55$}\\
\midrule
FedProto$$[\textcolor{violet}{5}]$$             & ~~$71.88$     & ~~$69.04$      & ~~$66.34$     & ~~$69.71$     & ~~$58.52$     & ~~$73.57$    & ~~\result{$68.18$}{$5.34$}\\
FedDistill$$[\textcolor{violet}{6}]$$                  & ~~$81.08$     & ~~$67.42$      & ~~$60.96$     & ~~$77.13$     & ~~$70.45$     & ~~$80.33$    & ~~\result{$72.90$}{$7.98$}\\
\midrule
\emph{FedEvPrompt}                     & ~~$81.02$     & ~~$71.83$      & ~~$84.15$     & ~~$79.02$     & ~~$68.18$     & ~~$79.35$    &~~\result{$\textbf{77.26}$}{$4.65$}       \\
\bottomrule
\end{tabular}
}
\end{table}

We finally conducted an ablation study to assess the impact of various prompting options and sharing strategies among nodes within the federation. It's worth noting that while prompt sharing may potentially compromise privacy guarantees, exploring its effectiveness compared to using private prompts and our proposed knowledge distillation approach is interesting.

To this end, we initially assessed the performance of a variant of FedAvg where only low-level \textit{b-prompts} are shared, gradually incorporating \textit{t-prompts} and knowledge distillation on the uncertainty-aware attention buffer. Furthermore, we examined the variant of the proposed prompting strategy using a single set of general prompts \textit{g-prompts} shared between nodes and coupled with our knowledge distillation method. Our findings, outlined in Table~\ref{tab:ablation}, underscore the significance of separate prompt learning, as evidenced by the subpar performance of the \textit{g-prompts} variants. Interestingly, sharing separate sets of \textit{b-prompts} and \textit{t-prompts} (first two rows of Tab.~\ref{tab:ablation}) proved less effective than keeping them private and employing knowledge distillation (best performance observed in the last two rows of the same table).
Moreover, we demonstrate that our strategy of incorporating attention maps based on uncertainty scores (as detailed in Sect.~\ref{sec:kd}) yields superior performance compared to random selection of buffer samples. 
These two last considerations highlight the informative contribution provided by the attention maps corresponding to the lowest uncertainty samples driving clients' models towards the most significant regions of skin lesion images.
\begin{table}
\centering
\caption{\textbf{Ablation study results} showing the impact of shared prompts and knowledge distillation (KD) on federated learning performance.}
\label{tab:ablation}
\rowcolors{2}{white}{gray!10}

\footnotesize{

\begin{tabular}{l|ccccccc} 
\toprule
                                        & \textbf{C1}   & \textbf{C2}  & \textbf{C3}  & \textbf{C4}  & \textbf{C5}  & \textbf{C6}  & Avg\\ 
\midrule

FedAvg $b_{prompts}$                    & ~~$50.00$     & ~~$70.04$    & ~~$70.29$    & ~~$79.61$    & ~~$71.59$    & ~~$79.21$    & ~~\result{$70.18$}{$7.56$}\\
\hskip 0.5em + $t_{prompts}$            & ~~$74.74$     & ~~$72.79$    & ~~$67.74$    & ~~$79.89$    & ~~$67.61$    & ~~$77.09$    & ~~\result{$73.31$}{$3.64$}\\
\hskip 1.5em + KD                       & ~~$73.02$     & ~~$71.41$    & ~~$76.80$    & ~~$81.74$    & ~~$67.61$    & ~~$79.25$    & ~~\result{$74.97$}{$3.94$}\\
\midrule
FedAvg $g_{prompts}$                & ~~$50.00$     & ~~$70.88$    & ~~$72.77$    & ~~$50.00$    & ~~$67.61$    & ~~$50.00$    & ~~\result{$60.21$}{$6.81$}\\
FedAvg $g_{prompts}$ + KD                & ~~$81.1$     & ~~$71.32$    & ~~$72.83$    & ~~$78.89$    & ~~$63.64$    & ~~$80.49$    & ~~\result{$74.71$}{$6.18$}\\
\midrule
KD\textsubscript{random}                   & ~~$78.12$     & ~~$69.40$    & ~~$76.80$    & ~~$80.30$    & ~~$68.75$    & ~~$77.65$    & ~~\result{$75.17$}{$3.66$} \\
KD\textsubscript{uncertainty} \textit{(Ours)}                     & ~~$81.02$     & ~~$71.83$    & ~~$84.15$    & ~~$79.02$    & ~~$68.18$    & ~~$79.35$    & ~~\result{$\textbf{77.26}$}{$4.65$}\\

\bottomrule

\end{tabular}
}
\end{table}

\section{Conclusion}
\label{sec:discussion}
This work introduces FedEvPrompt, a new federated learning approach tailored for skin lesion classification using the ISIC2019 dataset. Indeed, this dataset offers a realistic setting for evaluating federated learning methods, eliminating the need for simulated distributions.
FedEvPrompt seamlessly integrates evidential deep learning, prompt tuning, and knowledge distillation within a vision transformer architecture. Knowledge distillation on attention maps, in particular, ensures better privacy-preserving capabilities than parameter sharing. In addition to its superior performance in addressing data heterogeneity and privacy concerns, the employment of evidential learning offers enhanced model interpretability and uncertainty quantification, providing valuable insights for decision-making in medical image analysis. Balancing vacuity and dissonance \cite{josang2018uncertainty,guo2022survey} in buffer selection warrants further research to comprehensively understand underlying mechanisms.

\section*{Acknowledgements}We acknowledge the support of the PNRR ICSC National Research Centre for High Performance Computing, Big Data and Quantum Computing (CN00000013), under the NRRP MUR program funded by the NextGenerationEU. Rutger Hendrix is a PhD student enrolled in the National PhD in Artificial Intelligence, XXXVIII cycle BIS, course on Health and life sciences, organized by Università Campus Bio-Medico di Roma.

\bibliographystyle{unsrtnat}
\bibliography{biblio}  






\end{document}